\documentclass[sigconf,article, nonacm]{acmart}

\usepackage[inline]{enumitem}

\begin{document}

\title{ConceptFormer: Towards Efficient Use of Knowledge-Graph Embeddings in Large Language Models}

\author{Joel Barmettler}
\orcid{0009-0006-5118-7129}
\email{joel.barmettler@uzh.ch}
\affiliation{%
  \institution{University of Zurich}
  \city{Zurich}
  \country{Switzerland}
}

\author{Abraham Bernstein}
\orcid{0000-0002-0128-4602}
\email{bernstein@ifi.uzh.ch}
\affiliation{%
  \institution{University of Zurich}
  \city{Zurich}
  \country{Switzerland}
}

\author{Luca Rossetto}
\orcid{0000-0002-5389-9465}
\email{luca.rossetto@dcu.ie}
\affiliation{%
  \institution{Dublin City University}
  \city{Dublin}
  \country{Ireland}
}

\begin{abstract}

Retrieval Augmented Generation (RAG) has enjoyed increased attention in the recent past and recent advancements in Large Language Models (LLMs) have highlighted the importance of integrating world knowledge into these systems. 
Current RAG methodologies often modify the internal architecture of pre-trained language models (PLMs) or rely on textifying knowledge graphs (KGs), which is inefficient in terms of token usage.
This paper introduces \emph{ConceptFormer}, a new approach to augment LLMs with structured knowledge from KGs, such as Wikidata, without altering their internal structure or relying on textual input of KGs.
\emph{ConceptFormer} operates in the LLM embedding vector space, creating and injecting \emph{concept vectors} that encapsulate the information of the KG nodes directly.
Trained in conjunction with a frozen LLM, \emph{ConceptFormer} generates a comprehensive lookup table that maps KG nodes to their respective \emph{concept vectors}.
The approach aims to enhance the factual recall capabilities of LLMs by enabling them to process these \emph{concept vectors} natively, thus enriching them with structured world knowledge in an efficient and scalable manner.
Our experiments demonstrate that the addition of \emph{concept vectors} to GPT-2 0.1B substantially increases its factual recall ability (Hit@10) by up to 272\% when tested on sentences from Wikipedia and up to 348\% on synthetically generated sentences.
Even injecting only a single \emph{concept vector} into the prompt increases factual recall ability (Hit@10) by up to 213\% on Wikipedia sentences, significantly outperforming RAG with graph textification while consuming 130x fewer input tokens.

\end{abstract}



\maketitle

\section{Introduction}

\begin{figure}[t]
    \centering
    \vspace{5mm}
    \includegraphics[width=\linewidth]{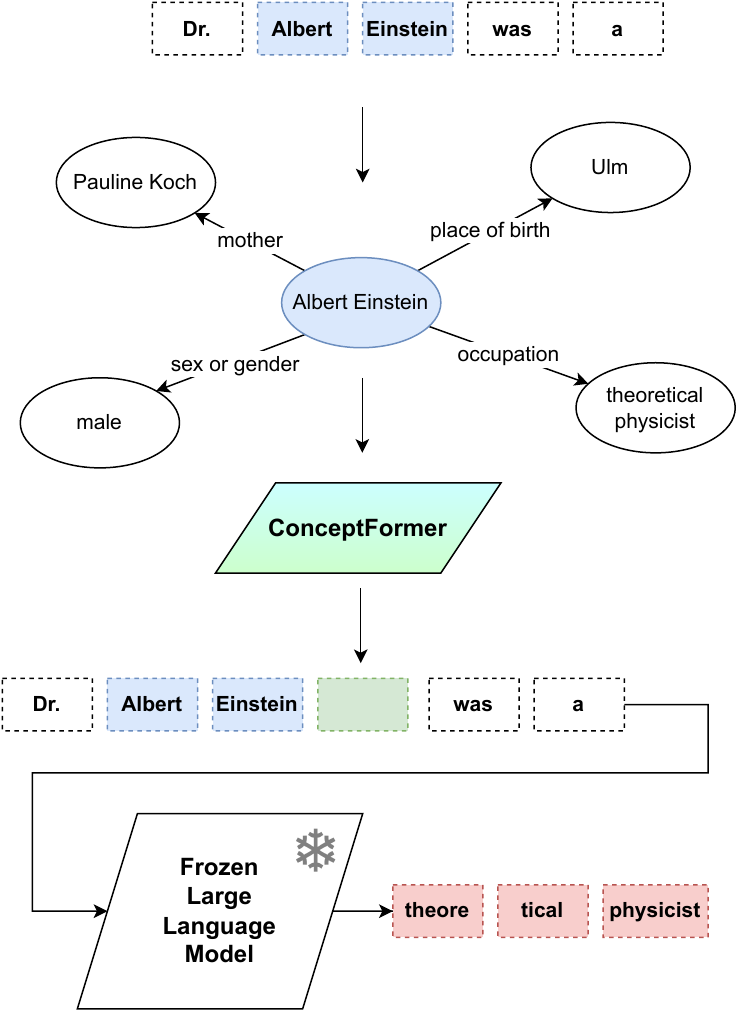}
    \caption{\emph{ConceptFormer} enhances a prompt by extending the embedding vectors from the original prompt with learned \emph{concept vectors}. Entity recognition and entity linking is used to detect an entity ``Albert Einstein'' (displayed in blue) in the original prompt and link it towards a large KG like Wikidata. \emph{ConceptFormer} creates a vector embedding for the detected entity that is compatible with the LLMs input embedding space. The resulting \emph{concept vector} (displayed in green) is shown to capture the essence of the entity far better than the original token embedding vectors alone, leading to more knowledgeable output generated by the LLM (displayed in red).}
    \Description{ConceptFormer enhances a prompt by extending the embedding vectors from the original prompt with learned concept vectors. Entity recognition and entity linking is used to detect an entity ``Albert Einstein'' in the original prompt and link it towards a large KG like Wikidata. ConceptFormer creates a vector embedding for the detected entity that is compatible with the LLMs input embedding space. The resulting concept vector (displayed in green) is shown to capture the essence of the entity far better than the original token embedding vectors alone, leading to more knowledgeable output generated by the LLM.}
    \label{fig:conceptformer-introduction}
\end{figure}

Large Language Models (LLMs) have demonstrated exceptional potential in various natural language processing tasks, including conversational agents, summarization, and information retrieval (IR). They are typically trained on large-scale, general-purpose (text) corpora and optimized via self-supervision objectives \citep{pretrained-llm-survey}. Through this pretraining, LLMs acquire substantial amounts of knowledge, which is implicitly stored within their model weights \citep{lima, knowledge-pack}. However, such implicit storage can lead to inefficient knowledge retrieval and risks of outdated, biased, or incomplete information, posing key obstacles for retrieval-augmented generation (RAG) and knowledge-intensive IR tasks \citep{rag, streamingqa, time-waits-no-one}.

A common approach to enhance knowledge retention in LLMs is through corpus curation or expansion, which has shown promise in reducing hallucinations \citep{creating-data, data-to-text}. But simply improving or enlarging the training corpus does not always resolve deeper IR challenges such as domain-specific knowledge retrieval or real-time updates \citep{llama2, textbooks}. Models still frequently hallucinate or omit key facts \citep{bertyet, factscore}, especially in specialized domains where domain shifts are inevitable \citep{chatgpt-robustness}.
Research has shown that growing the parameter count of LLMs can improve factual recall \citep{factuality}, but even large models like ChatGPT still struggle to accurately retrieve and articulate structured knowledge, e.g., from DBPedia \citep{ontological-knowledge, head-to-tail}. Such difficulties reflect broader IR concerns: a system's inability to efficiently retrieve relevant concepts hampers downstream tasks like question answering, recommendation, and conversational search.

To address these challenges in knowledge-intensive scenarios, Knowledge Graphs (KGs)~\citep{kg-survey, never-ending-learning} have emerged as a valuable structured resource. KGs---such as Wikidata \citep{wikidata}---capture accurate, up-to-date, and domain-rich factual information. Making this graph-based knowledge accessible to LLMs is a longstanding goal in IR research, especially for retrieval-augmented generation and question answering \citep{knowledge-enhance-survey, domain-specific-kg}. Popular RAG-focused approaches typically rely on ``\emph{textification}'' of graph edges or nodes \citep{talklikegraph, graphrag}, concatenating them into the LLM prompt. While effective, this strategy consumes large portions of the model’s context window and can introduce noise \citep{knowledge-noise}.

Consider a conversational IR system that must answer a user’s query about a niche topic (e.g., ``What type of mustache did \emph{Albert Einstein} famously sport?''). If the system is forced to feed hundreds of tokens of textified KG neighbors into the LLM prompt for every entity, it risks hitting context limits or drowning in irrelevant details. An alternative is to \emph{embed} this knowledge in a more compact form, seamlessly integrating it with the user query inside the LLM’s embedding space. Such a solution would not only save tokens but also yield more targeted retrieval results, echoing a core IR theme: balancing retrieval accuracy with efficiency.

In this paper, we introduce \emph{ConceptFormer}, a novel, token-efficient way to integrate KGs into any pre-trained LLM without modifying its internal architecture or retraining its core parameters. Rather than textifying knowledge, \emph{ConceptFormer} injects \emph{concept vectors} derived from the KG directly into the LLM’s input embedding space. 
Figure~\ref{fig:conceptformer-introduction} illustrates how entity recognition links a text mention (``Albert Einstein'') to its corresponding Wikidata node, and \emph{ConceptFormer} learns to produce a small set of dense \emph{concept vectors} that augment or replace the naive token embeddings. These vectors \emph{natively} encode the most relevant graph-neighborhood information, enabling the LLM to integrate structured knowledge with minimal context overhead.

The contributions of this paper are threefold:
\begin{enumerate*}
    \item We present \emph{ConceptFormer}, a flexible mechanism to embed KG nodes into the LLM prompt space, enabling IR scenarios that demand large-scale or domain-specific factual retrieval without altering the LLM architecture.
    \item We introduce new datasets---\emph{Tri-REx}, \emph{T-REx Bite}, and \emph{T-REx Star}---specifically designed to evaluate next-token and factual-recall tasks.
    These datasets facilitate measuring how well LLMs retrieve, re-rank, and generate entity-level facts.
    \item We empirically demonstrate that \emph{ConceptFormer} achieves \textbf{up to 348\% improvement} in factual recall (Hit@10) on synthetic sentences, and \textbf{up to 272\% improvement} on Wikipedia-based sentences, compared to a GPT-2 0.1B baseline. Remarkably, even a \emph{single concept vector} yields a competitive recall boost while consuming $\mathbf{130\times}$ fewer tokens than text-based RAG.
\end{enumerate*}
We make all datasets, as well as our implementation\footnote{\url{https://github.com/joelbarmettlerUZH/ConceptFormer}} and pre-trained models~\cite{zenodo_models} freely available for download.

Whereas most KG-enhanced LLM research focuses on the \emph{text representation of knowledge}~\citep{colake, kbert, dkplm}, we focus on \emph{vector-based injection} that compresses graph information directly within the LLM’s input space. We deliberately chose GPT-2 0.1B~\citep{gpt2} for our experiments due to its comparatively small size and simple architecture. However, the technique extends to larger LMs with minimal adaptation. 

We structure the remainder of the paper as follows:  
    Section \ref{chapter:related-work} surveys related methods of retrieval-augmented generation, prompt-tuning, and knowledge-graph enhancements.
    Section \ref{chapter:datasets} describes the newly introduced datasets and their alignment with IR tasks.
    Section \ref{chapter:method} details the \emph{ConceptFormer} architecture, its training process, and how it injects compressed KG embeddings into prompts.
    Section \ref{chapter:experiments} presents extensive experimental results, including comparisons to RAG baselines, analysis of token usage, and a question-answering scenario.
    Section \ref{chapter:conclusion} concludes with implications for ret\-rieval-based LLM applications and future work.

By demonstrating a \emph{vector-centric} approach to knowledge injection that preserves the LLM’s original weights, we hope to inspire further IR research on KG-based efficient, up-to-date retrieval for knowledge-intensive generation tasks.


\section{Related Work}
\label{chapter:related-work}

\emph{ConceptFormer} intersects several research areas that are central to information retrieval (IR) in the presence of Large Language Models (LLMs). We situate our work among (i) retrieval-augmented generation (RAG), (ii) KG-enhanced LLMs, (iii) token compression or ``gisting,'' (iv) prompt tuning, and (v) pseudo-word embeddings for multimodal or specialized concepts.

\paragraph{Retrieval-Augmented Generation (RAG)}

Retrieval-augmented generation (RAG) \citep{rag,ragsurvey} enhances LLM-based text generation by retrieving relevant external data before or during text generation, thus grounding model outputs in factual information from structured or unstructured sources. In conventional IR pipelines, this approach is reminiscent of query expansion or contextualization: the user query (or partial text) is augmented with retrieved documents, enabling the LLM to generate content that more accurately reflects the retrieved evidence.

A straightforward yet potent variant is \emph{graph textification}, wherein a knowledge graph (KG) is linearized into textual templates \citep{talklikegraph}. For instance, every subject--predicate--object triple can be converted into a text phrase and concatenated with the user query. Works like \citet{promptenrichment} and \citet{graphrag} rely on these textual expansions to harness structured knowledge in an LLM’s generation process. While effective, this strategy often imposes heavy token overhead, straining context windows and introducing potential noise when many graph edges are involved.

By contrast, \emph{ConceptFormer} avoids textifying knowledge. Rather than inserting hundreds of tokens describing a KG neighborhood, it transforms the KG node (and its adjacency information) into a small set of dense, learned vectors directly injected into the LLM’s embedding space. This significantly reduces context usage while preserving the ability to retrieve and integrate graph-based evidence.

\paragraph{KG-Enhanced LLMs for IR}
\label{sec:kg-llm}

Beyond RAG, multiple lines of research aim to integrate KG information into an LLM’s representation. Early works frequently required extensive architecture modifications or additional training heads. For example:
    \textit{DKPLM} \citep{dkplm} dynamically updates language models with knowledge extraction and pseudo-token injections, altering model layers and unfreezing part of the network.
    \textit{CoLAKE} \citep{colake} builds a hybrid word-knowledge graph and modifies both the embedding and encoder layers of the Transformer, training an LLM from scratch.
    \textit{K-Bert} \citep{kbert}, \textit{KnowBert} \citep{knowbert}, and \textit{KP-PLM} \citep{kpplm} each propose new attention or injection layers, partially or fully overriding a model’s internal architecture.
    \textit{K-Adapter} \citep{kadapter} appends adapters specialized in certain types of knowledge.
%
While these methods effectively enhance LLMs with structured knowledge, most are tailored to encoder-only models (e.g., BERT) or require partial fine-tuning of the LLM. This complicates their usage in RAG deployments, where one may prefer to keep large-scale pretrained models intact to preserve existing capabilities.

\emph{ConceptFormer} aligns with these methods in its goal — making KG information accessible to the LLM — but preserves the fundamental architecture of a decoder-only LLM by operating strictly at the \emph{input-embedding} level. In RAG scenarios with minimal inference memory or a preference for plug-and-play modules, \emph{ConceptFormer} can be combined with a standard LLM in a non-invasive way. Furthermore, its \emph{concept vectors} can be \emph{precomputed} and stored for fast retrieval, eliminating inference overhead of reconstructing graph neighbors on the fly.

\paragraph{Gist Tuning and Prompt Compression}
\label{sec:gist}

Gist tuning \citep{gist} condenses lengthy prompts into more compact ``gist tokens.'' Similar to knowledge textification, large prompt expansions can degrade performance or exceed context limits in tasks that rely on extensive background passages. By training a compressor to produce a short sequence of learnable tokens, gist tuning reduces the number of tokens needed, often maintaining high-quality generation.

The idea of compressing large bodies of text into a minimal embedding resonates with \emph{ConceptFormer}, where entire KG neighborhoods are expressed as a few \emph{concept vectors}. Instead of elaborating text expansions, \emph{ConceptFormer} forms a high-level ``gist'' of an entity’s local subgraph. This is especially beneficial in retrieval-based LLM use cases: the context window is left free for user queries or additional data, yet the LLM still has access to structured knowledge.

\paragraph{Prompt Tuning and Continuous Prompts.}
\label{sec:prompt-tuning}

Prompt tuning approaches \citep{gptunderstands,prefixtuning,howtoask} freeze most or all of an LLM’s parameters, introducing continuous embeddings (``prefix vectors'' or ``soft prompts'') that steer the model’s behavior. This differs from typical fine-tuning, which updates the model’s internal weights and can lead to catastrophic forgetting of original capabilities. For IR tasks involving many domain-specific expansions or diverse subtasks, preserving the LLM’s core weights can be advantageous.

\emph{ConceptFormer} extends this paradigm: rather than encoding a generic style or instruction, it injects \emph{entity-centric knowledge}. Each KG node is accompanied by one or more learned embedding vectors that store local relational context. This is akin to ``soft prompts'' but specifically aimed at knowledge injection in a RAG setting.

\paragraph{Pseudo-Words for Specialized Concepts.}
\label{sec:pseudo-words}

A related vein of research addresses new concept acquisition in LLMs by inserting ``pseudo words'' into the model’s input space. In multimodal or domain-adaptation contexts, these pseudo tokens can represent, for instance, novel visual concepts \citep{oneword,frozen} or domain-specific terms. The idea is to attach new semantics to an otherwise unused token embedding, enabling the model to integrate or reference the concept during generation.

\emph{ConceptFormer} similarly represents new concepts (KG entities) via dense vectors, though it differs by focusing on subgraph-level knowledge, not just an image or a single domain label. By mapping an entire entity neighborhood into a few vectors, it preserves the relational structure in a compressed embedding. This synergy of local graph information and token-space injection allows LLMs to more accurately recall factual connections relevant for IR tasks such as entity-centric question answering, knowledge-grounded summarization, or domain-targeted retrieval.


\section{Datasets}
\label{chapter:datasets}

Information retrieval tasks often require evaluating whether a model can retrieve the correct pieces of knowledge and ground its responses in accurate facts. Large Language Models generally learn from massive corpora that blend linguistic and factual content (e.g., Wikipedia), but such corpora alone are often suboptimal when the goal is to infuse \emph{new} facts or systematically measure how effectively a model recalls or generates factual information in a controlled manner.

To address this issuem we introduce three datasets—\emph{T-REx Bite}, \emph{Tri-REx}, and \emph{T-REx Star}—that build upon the foundation of the \emph{T-REx} dataset~\citep{trex}. While T-REx links Wikipedia sentences to Wikidata triples, it was not designed specifically for next-token prediction. By contrast, our datasets provide explicit structures that facilitate both \emph{knowledge injection} and \emph{retrieval-based} LLM evaluations, addressing key limitations of T-REx and its LAMA-based extensions \citep{lama, lamacontext}. These limitations include the assumption of single-token objects, uneven coverage of knowledge, and inadequate alignment with the needs of decoder-only LLMs often used in IR settings.

\subsection{T-REx Bite}
\label{sec:trex-bite}

\emph{T-REx Bite}~\citep{trexbite_zenodo} adapts T-REx to the next-token-prediction paradigm by ensuring that, in each text snippet, the subject appears before the object. This alignment mimics real-world scenarios in which a model sees partial information (the subject and some context) and must then predict or ``retrieve'' the missing object. To keep snippets manageable within the limited context windows of smaller language models (such as GPT-2), each snippet is capped at 512 characters. We further require that the snippet explicitly mentions both subject and object, does not start a new sentence at the object, and is linked to the corresponding subgraph in T-REx Star (see Section~\ref{sec:trex-star}). 

By applying these constraints, we obtain about 6.4 million short ``bites'' for training, 0.92 million for testing, and 0.75 million for validation. Each bite is a compact piece of Wikipedia text that retains the original clarity and diversity of T-REx but is tailored to ensure a direct subject--object alignment. This structure lets researchers readily evaluate how well an LLM completes the object token(s) given the preceding subject. The dataset naturally accommodates multi-token objects under modern sub-word tokenization, removing the single-token assumptions of LAMA-like methods.

\subsection{Tri-REx}
\label{sec:tri-rex}

\begin{figure}[t]
    \includegraphics[width=\linewidth]{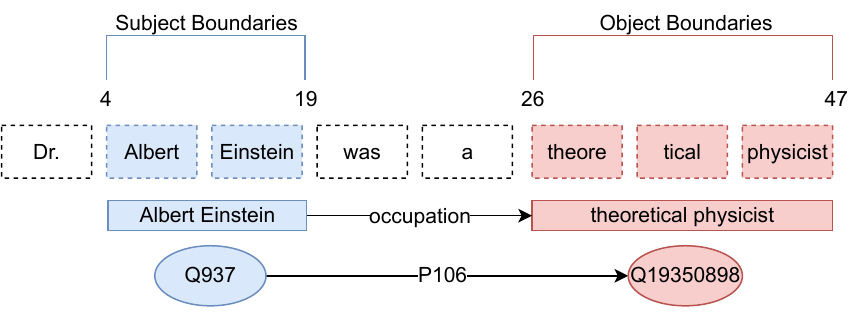}
    \caption{Example datapoint from \emph{Tri-REx} (Synthetic) Dataset. The datapoint consists of the main sentence(s), information about the mentioned Wikidata triple, as well as boundary indications of the entity label locations within the sentences(s).}
    \Description{Example datapoint from the synthetic Tri-REx Dataset. The datapoint consists of the main sentence(s), information about the mentioned Wikidata triple, as well as boundary indications of the entity label locations within the sentences(s).}
    \label{fig:trirex-dataset-showcase}
\end{figure}

Although T-REx Bite is useful, it still relies on Wikipedia text that many models may have partially seen during their pretraining phase. To create a scenario free from such potential contamination, we introduce \emph{Tri-REx}~\cite{trirex_zenodo}. Instead of extracting text from Wikipedia, Tri-REx synthesizes short subject--predicate--object sentences using Mistral 7B \citep{mistral7b} in a few-shot prompting fashion. For example, a triple \textit{(Albert Einstein, facial hair, walrus moustache)} might generate ``Dr.\ Albert Einstein wore a bushy walrus moustache.'' Each generated sentence is automatically filtered for coherence, correct mention of both subject and object, and accurate preservation of the S-P-O relationships, resulting in high-quality synthetic data. An illustration of a datapoint from \emph{Tri-REx} is given in Figure~\ref{fig:trirex-dataset-showcase}.

Tri-REx comprises 21.5 million training sentences, 0.9 million test sentences, and 1.7 million validation sentences, each of which is typically under 30 tokens. This collection stands out because it is intentionally \emph{free of pretraining overlap}: models cannot simply rely on memorized Wikipedia text. Instead, they must learn or leverage newly provided knowledge sources (e.g., \emph{concept vectors} from ConceptFormer) to recover the correct object tokens during next-token prediction. Researchers can thus verify whether a RAG technique or knowledge-injection approach genuinely conveys facts to a model, rather than merely triggering recall of memorized text. 

\subsection{T-REx Star}
\label{sec:trex-star}

While T-REx Bite and Tri-REx focus on textual input, they do not explicitly embed the wider \emph{graph structure} that connects subject entities to their neighbors. \emph{T-REx Star}~\cite{trexstar_zenodo} fills this gap by providing star-topology subgraphs from Wikidata for each entity that appears as a subject in T-REx. Each entity’s local subgraph is represented in JSON format, includes up to 100 neighbors ranked by PageRank~\citep{pagerank}, and stores both node (Q-ID, English label, PageRank) and edge (P-ID, relation label) metadata. The JSON structure is easily loaded into tools such as NetworkX~\citep{networkx}, enabling further graph-based processing or embedding.

Crucially, T-REx Star aligns with T-REx Bite and Tri-REx by using a consistent partitioning scheme. Every entity that serves as a \emph{subject} in one of the three datasets appears in exactly one split (train, validation, or test). Entities may nonetheless appear as \emph{objects} in multiple splits if they are neighbors of different subjects. This consistency is important for fair comparisons of LLM performance across training and evaluation sets.

\subsection{Relevance and Utility for IR}

By design, these three datasets complement one another and support a broad range of IR-driven studies:

\textbf{T-REx Bite} preserves the naturalness of original Wikipedia text. It is suitable for real-world RAG setups where subject--object pairs appear in an authentic linguistic context. Evaluating next-token prediction on T-REx Bite reveals how effectively the model can fill in factual objects without exceeding realistic context limits.

\textbf{Tri-REx} sheds potential contamination by synthesizing new S-P-O sentences for each entity–neighbor pair. It thus focuses the learning and evaluation strictly on externally provided facts, an ideal setup for measuring how well knowledge-injection approaches can impart new or domain-specific information into an LLM.

\textbf{T-REx Star} explicitly includes the surrounding Wikidata structure for each entity, enabling research into how local graph neighborhoods can inform retrieval-augmented LLM usage. Instead of textifying these subgraphs, systems like \emph{ConceptFormer} can inject them efficiently as a small set of learned embeddings.

All three datasets share consistent train, validation, and test splits based on subject partitioning, ensuring controlled experimentation in knowledge retrieval and generation tasks. They are publicly released alongside this work to foster new approaches that integrate structured knowledge into LLMs without over-relying on textual expansions. By offering both natural and synthetic data, with or without explicit KG neighborhoods, these resources aim to push the boundaries of how IR systems can leverage knowledge injection for next-token and factual-recall evaluations.


\section{Method}
\label{chapter:method}

\emph{ConceptFormer} is designed to inject compact, graph-based knowledge into a Large Language Model (LLM) without altering the LLM’s internal architecture. In the context of information retrieval, this design choice is essential: large pretrained models are often deployed ``as is'' due to computational constraints or fear of catastrophic forgetting. By operating purely at the \textit{input embedding} level, \emph{ConceptFormer} seamlessly integrates with most decoder-only LLMs and can be flexibly applied in retrieval-augmented generation, entity-centric search, or domain-specific IR pipelines.

\begin{figure}[t]
    \centering
    \includegraphics[width=\linewidth]{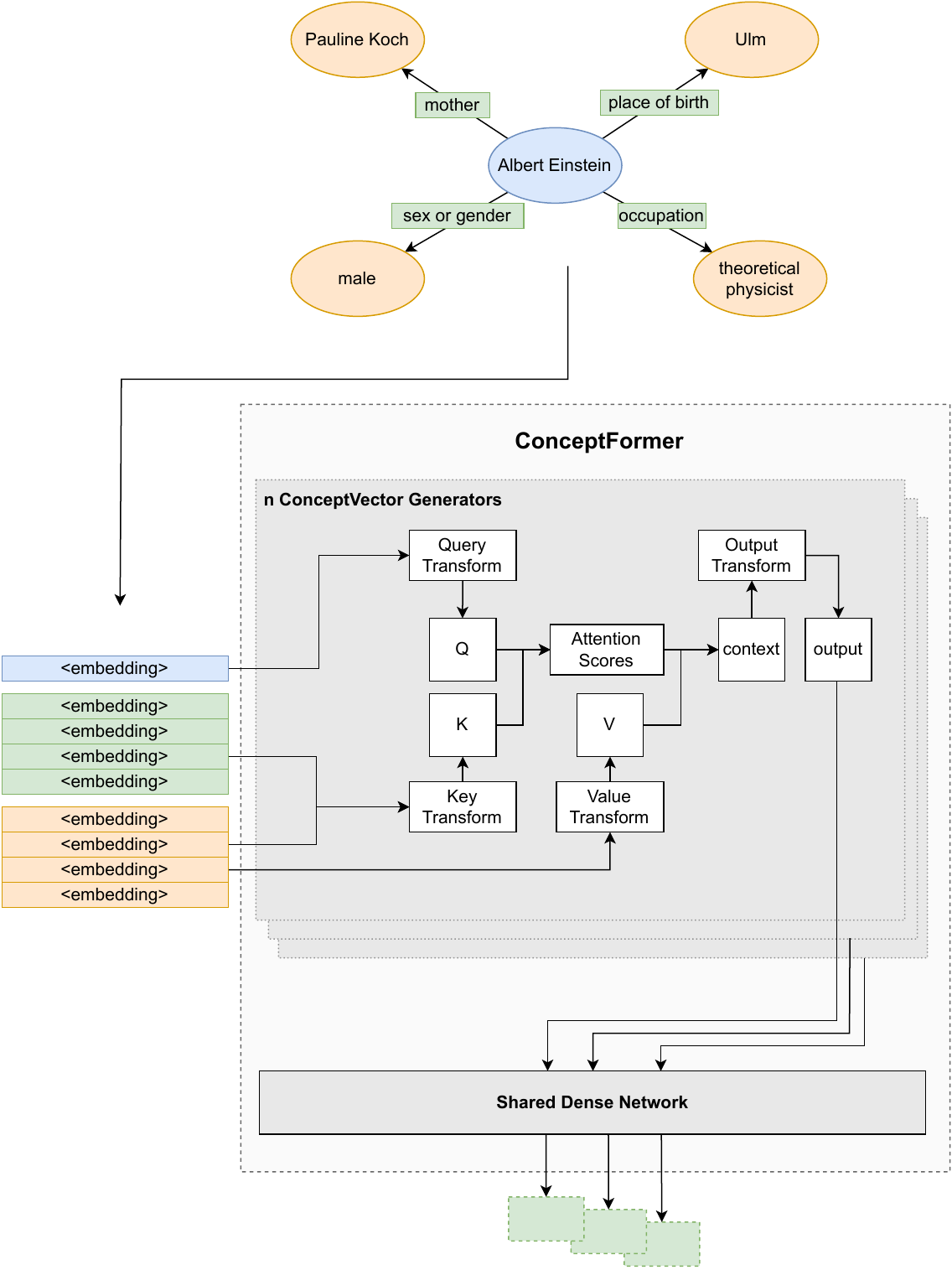}
    \caption{The input of the \emph{ConceptFormer} are three matrices, representing the central node, neighbouring nodes, and connecting edges. These embeddings can be generated with numerous text-embedding mechanisms. In our work, we generated the node and edge embeddings by simply forwarding their label through an LLM and averaged the last hidden layer. \emph{ConceptFormer} trains multiple, parallel, and fully independent \emph{concept vector generator blocks}, each implementing an attention mechanism in which the central node becomes the query Q, the concatenated neighbouring nodes and corresponding edges become the key K, and the neighbouring nodes become the values V. Finally, a shared dense network transforms the output of each \emph{concept vector generator block} into the input embedding space of the LLM.}
    \Description{The input of the ConceptFormer are three matrices, representing the central node, neighbouring nodes, and connecting edges. These embeddings can be generated with numerous text-embedding mechanisms. In our work, we generated the node and edge embeddings by simply forwarding their label through an LLM and averaged the last hidden layer. ConceptFormer trains multiple, parallel, and fully independent concept vector generator blocks, each implementing an attention mechanism in which the central node becomes the query Q, the concatenated neighbouring nodes and corresponding edges become the key K, and the neighbouring nodes become the values V. Finally, a shared dense network transforms the output of each concept vector generator block into the input embedding space of the LLM.}
    \label{fig:architecture}
\end{figure}

At a high level, \emph{ConceptFormer} can be viewed as a \emph{modular knowledge injector}:
\begin{enumerate}
    \item \textbf{Entity Detection and Linking:} Given a user query or partial text (e.g., ``Albert Einstein was a\dots''), an off-the-shelf Named Entity Recognition (NER) and entity linker identifies ``Albert Einstein'' and retrieves its corresponding node ID (Q\_ID) in Wikidata or any other KG.
    \item \textbf{Subgraph Extraction:} A star topology subgraph around the entity is fetched, containing its immediate neighbors and the edges (predicates) connecting them. In our experiments (\S\ref{chapter:datasets}), subgraphs are pre-extracted from \emph{T-REx Star}. 
    \item \textbf{\emph{ConceptFormer} Vector Generation:} The extracted node and edge embeddings are fed into \emph{ConceptFormer}, yielding \emph{concept vectors} that capture the local neighborhood. 
    \item \textbf{Prompt Extension:} These \emph{concept vectors} are appended to the existing input embeddings of the subject, forming a richer representation that the LLM processes natively (Figure~\ref{fig:architecture}). 
\end{enumerate}

This procedure allows an LLM to incorporate structured graph knowledge with minimal prompt overhead. Unlike text-based RAG approaches that concatenate large textual expansions (sometimes hundreds of tokens), \emph{ConceptFormer} only inserts $n$ vectors per entity---where $n$ is typically far smaller (e.g., 1 to 20 vectors). As a result, it significantly reduces context consumption, freeing up tokens for user text or system instructions in real-world RAG scenarios.

\subsection{Architecture of ConceptFormer}
\label{subsec:conceptformer-architecture}

\emph{ConceptFormer} takes as input a \emph{star topology subgraph} anchored on a central entity, which we denote as $C$. That entity has $m$ neighbors $N_1,\dots,N_m$ and corresponding edge (predicate) embeddings $E_1,\dots,E_m$. Formally, each neighbor $N_i$ is a fixed-dimensional vector (e.g., 768\,D) representing the neighbor’s label or textual description, and each edge $E_i$ is similarly embedded. These embeddings can be derived using off-the-shelf text embedding models (e.g., GPT-2, Word2Vec) or specialized methods like TransE \citep{transe} or PBG \citep{pbg}.

In line with \citet{modularmuldimodal}, we separate (i) the \textbf{node/edge embedding} step from (ii) the \textbf{alignment and compression} step. The former yields $C, N, E$ with dimension $\text{dim}_i$, and the latter is where \emph{ConceptFormer}’s trainable parameters reside.

\emph{ConceptFormer} implements $n$ parallel concept vector generator blocks. Each of these generators learns a Key ($K_n$), Query ($Q_n$), Value ($V_n$), and Output ($O_n$) transformation through linear layers with weights $W^Q_n$, $W^k_n$, $W^V_n$, and $W^O_n$. The output of each concept attention layer is an intermediate \emph{concept vector}  \((1, \text{dim}_i)\), see Equations \ref{eq:1}-\ref{eq:4}.

Each $O_n$ is then processed through an output transformation shared by all concept attention generators to produce the final \emph{concept vector}s, see Equation \ref{eq:conceptvector}.
This transformation is critical for converting \emph{concept vectors} into a format compatible with the LLM's input space, a technique also employed in other papers to bridge the gap between graph representations like TransE and LLM text embeddings \citep{commonsense-generation}.
The output dimension $\text{dim}_o$ of this layer is defined by the input token embedding used by the LLM.
Its hidden dimension is a free parameter, which we set to 1228 in our experiments.

\begin{align}
Q_n &= C W^Q_n \label{eq:1}\\
K_n &= N W^K_n + E \\
V_n &= N W^V_n \\
O_n &=  \left(softmax \left( \frac{Q_n {K_n}^T}{\sqrt{\text{dim}_i}} \right) V_n \right) W^O_n 
\label{eq:4}
\end{align}
\begin{equation}
    \text{\emph{concept vector}}_n = LeakyReLu(O_n W^P_1)W^P_2
    \label{eq:conceptvector}
\end{equation}

\emph{ConceptFormer} can be fine-tuned to adapt to a new LLM, making it a versatile tool that can be integrated with various language models.
The output of \emph{ConceptFormer} is a set of $n$ \emph{concept vectors}, forming a matrix of size $(n, \text{dim}_o)$. This matrix represents the transformed knowledge from the input subgraph, ready to be fed into the LLM for enriched language generation.

\subsection{Training Objectives and Stages}

\label{subsubsec:pretrain}

We first train \emph{ConceptFormer} on synthetic sentences from \emph{Tri-REx} (\S\ref{chapter:datasets}), which are deliberately \emph{unseen} by the base LLM. This ensures that the LLM cannot trivially predict the correct object from memorized textual patterns. In other words, it must rely on the knowledge embedded in the \emph{concept vectors}.
For each synthetic sentence of the form ``\texttt{[Subject] ... [Predicate] ... [Object]}'':
\begin{enumerate}
    \item We truncate the sentence right before the \texttt{[Object]}.
    \item We retrieve the star subgraph of the \texttt{[Subject]} from \emph{T-REx Star}, generate $n$ concept vectors using \emph{ConceptFormer}, and insert them after the original \texttt{[Subject]} text embedding vectors.
    \item The LLM is tasked with next-token prediction of the \texttt{[Object]}, minimizing the cross-entropy loss over the ground-truth tokens.
\end{enumerate}
Since the LLM is frozen, gradient updates only affect \emph{ConceptFormer} parameters. Over repeated batches, \emph{ConceptFormer} learns to encode the subgraph in a way that helps the LLM correctly predict the object tokens, even though the LLM itself has not been fine-tuned on these unseen facts.

\label{subsubsec:finetune}

After pre-training, we further refine \emph{ConceptFormer} using real-world text from \emph{T-Rex Bite}, ensuring it transfers from the synthetic domain to more natural, context-rich sentences. This second stage is crucial to avoid an over-simplified reliance on “triplet schemas” alone. Concretely:
\begin{itemize}
    \item We apply the same next-token prediction objective on truncated real sentences from Wikipedia, referencing the same entity subgraphs.
    \item The \emph{concept vectors} must adapt to the varied, noisier style of genuine Wikipedia text rather than the neatly structured synthetic statements.
\end{itemize}
This two-stage training (synthetic $\to$ real text) has proved effective in our experiments (\S\ref{chapter:experiments}), allowing \emph{ConceptFormer} to handle both contrived and authentic RAG use cases—ranging from short fact queries to more elaborate, context-dependent queries.

\subsection{Implementation Details and Training Hyperparameters}

In our experiments, each concept vector generator block uses linear transformations of dimension $\text{dim}_i \rightarrow \text{dim}_i$, with $\text{dim}_i$ typically 768. The subsequent MLP has one hidden layer of size 1228, using a LeakyReLU activation. We vary the number of parallel blocks $n \in \{1,2,3,4,5,10,15,20\}$, balancing performance and memory overhead.
We adopt AdamW with weight decay of $0.01$ and a constant learning rate, typically $6\times 10^{-5}$, determined via grid or Bayesian hyperparameter search. We freeze all LLM weights, ensuring stable optimization for \emph{ConceptFormer} alone.

We batch multiple subgraphs and partial sentences, limiting each to a maximum token length (e.g., 512) for GPT-2 compatibility. We early-stop if validation loss does not improve after one full epoch in either the synthetic or real-data stage, preventing overfitting to narrower patterns.
Once training is complete, \emph{ConceptFormer} can be used in two ways:
\begin{enumerate}
    \item \textbf{On-the-Fly Generation:} For each entity mention encountered during inference, dynamically retrieve its subgraph and run it through \emph{ConceptFormer} to produce \emph{concept vectors}. This supports real-time updates if the KG changes frequently but requires additional inference computation.
    \item \textbf{Precomputed Lookup Table:} For static KGs like a fixed snapshot of Wikidata, generate \emph{concept vectors} for all entities offline and store them in a giant key--value map. During inference, simply fetch the relevant \emph{concept vectors} for each detected entity, incurring near-zero overhead. 
\end{enumerate}

The second approach is often attractive in IR systems with stable or slow-changing knowledge bases, as it avoids re-running \emph{ConceptFormer} repeatedly for the same entities.


\section{Experiments}
\label{chapter:experiments}

In this section, we investigate how well a Large Language Model can recall facts from a knowledge graph, with a particular focus on Wikidata. We frame our experiments from an IR perspective: the model is presented with partial text (akin to a user query plus some context) and must \emph{retrieve or recall} the factual object entity. By isolating knowledge-intensive tasks, we aim to show how \emph{ConceptFormer} enables compact knowledge integration without exceeding context budgets or re-training the LLM.

\subsection{Evaluation Paradigm and Metrics}
\label{subsec:evaluation-setup}

We base our experiments on the \emph{Tri-REx} and \emph{T-REx Bite} datasets introduced in Section \ref{chapter:datasets}. Both sets consist of sentences truncated just before the mention of a target object entity (e.g., ``Albert Einstein was a\dots''), and the LLM is tasked with predicting the next tokens corresponding to the true entity label. 

To quantify recall, we adopt the widely used Hit@k metric. For an object label split into $T$ tokens, we record the rank $(r_t)$ of each token $t$ in the model’s output logits. The sequence rank is taken as $r = \max\{r_1,\ldots,r_T\}$, and it counts as a ``hit'' if $r \le k$ (i.e., all tokens appear in the top-$k$ predictions at their respective timesteps). This approach is robust to multi-token entities, a common challenge in IR tasks involving named entities (``New York Times'' vs.\ ``NYT'').

Restricting to Hit@1 (top-1 predictions) may understate the model’s factual knowledge, since many facts can be phrased in multiple correct ways. For retrieval tasks, we are interested in whether the ground-truth can appear \emph{at all} among the top few candidates. Hence we primarily highlight Hit@10, though we report Hit@1 and Hit@5 for completeness. A large gap between Hit@1 and Hit@10 may also reveal potential user experience differences in real-world IR.

\subsection{Baseline Evaluation}
\label{sec:eval-baseline}

We begin by establishing baseline performance for six LLMs of varying sizes and architectures:
\begin{itemize}
    \item \textbf{GPT-2 0.1B, 0.3B, 0.7B, 1.5B} \citep{gpt2}
    \item \textbf{LLaMA-2 3B, 7B} \citep{llama2}
\end{itemize}
Each model is evaluated \emph{as is}, without any knowledge injection. We measure Hit@1, Hit@5, and Hit@10 on the test splits of \emph{Tri-REx} and \emph{T-REx Bite}.

Table~\ref{tab:results} (rows without RAG or CF) shows that larger models consistently outperform their smaller counterparts across both datasets, reflecting well-known scaling trends \citep{factuality}. The gap between \emph{Tri-REx} and \emph{T-REx Bite} is particularly striking at Hit@1: performance on the Wikipedia-based \emph{T-REx Bite} is often 5--10$\times$ higher than on the synthetic \emph{Tri-REx}, suggesting that the model leverages memorized textual patterns from its pretraining corpus. This phenomenon highlights a key challenge: generalizing beyond memorized facts to new or rare knowledge.

Moreover, LLaMA-2 variants (3B and 7B) display notably better results than GPT-2 models of even larger parameter counts (e.g., GPT-2 1.5B), implying that architecture and training methodology can greatly influence factual recall in IR tasks.

\begin{table}[t]
    \centering
    \caption{Percentage of correctly predicted entities per dataset of different models using no augmentation, textified graph injection, or using different ConceptFormer-\emph{n} variants, producing between 1 and 20 concept vectors.}
    \resizebox{\linewidth}{!}{
    \begin{tabular}{lcccccc}
        \toprule
        & \multicolumn{3}{c}{\textbf{Tri-REx}} & \multicolumn{3}{c}{\textbf{T-Rex Bite}} \\
        \cmidrule(lr){2-4} \cmidrule(lr){5-7}
        \textbf{Model} & \textbf{H@1} & \textbf{H@5} & \textbf{H@10} & \textbf{H@1} & \textbf{H@5} & \textbf{H@10} \\
        \midrule
        LLaMA-2 7B  & 4.1\% & 17.5\% & 24.5\% & 39.3\% & 65.3\% & 73.0\% \\
        LLaMA-2 3B  & 4.3\% & 16.4\% & 22.9\% & 34.8\% & 59.5\% & 67.5\% \\
        GPT-2 1.5B  & 1.7\% &  8.8\% & 12.7\% & 22.9\% &  43.8\% & 51.8\% \\
        GPT-2 0.7B  & 1.6\% &  8.8\% & 12.4\% & 19.7\% &  39.3\% & 47.1\% \\
        GPT-2 0.3B  & 0.9\% &  7.0\% &  10.3\% & 16.3\% &  36.2\% &  44.2\% \\
        GPT-2 0.1B & 1.3\% &  5.8\% &  8.5\% & 4.7\% &  14.3\% &  19.5\% \\
        \midrule
        LLaMA-2 7B + RAG  & 25.6\% & 61.0\% & 72.2\% & 55.3\% & 85.1\% & 90.6\% \\
        LLaMA-2 3B + RAG  & 28.4\% & 61.5\% & 71.7\% & 52.0\% & 82.3\% & 88.4\% \\
        GPT-2 1.5B + RAG  & 26.0\% &  54.2\% & 63.8\% & 39.6\% &  70.3\% & 78.3\% \\
        GPT-2 0.7B + RAG  & 20.9\% &  48.9\% & 59.1\% & 30.9\% &  60.0\% & 69.3\% \\
        GPT-2 0.3B + RAG  & 21.5\% &  50.7\% &  59.9\% & 30.5\% &  62.3\% &  71.4\% \\
        GPT-2 0.1B + RAG & 8.3\% & 32.1\% & 41.3\% & 6.6\% &  23.8\% &  32.4\% \\
        \midrule
        GPT-2 0.1B + CF-20  & 16.8\% & 31.2\% & 36.9\% & 46.1\% & 65.6\% & 70.7\% \\
        GPT-2 0.1B + CF-15  & 16.2\% & 32.3\% & 38.2\% & 46.7\% & 67.1\% & 72.5\% \\
        GPT-2 0.1B + CF-10  & 15.7\% & 31.6\% & 37.8\% & 46.4\% & 67.2\% & 72.9\% \\
        GPT-2 0.1B + CF-5  & 13.6\% & 28.6\% & 35.1\% & 42.1\% & 63.3\% & 69.3\% \\
        GPT-2 0.1B + CF-4  & 12.7\% &  27.1\% & 33.7\% & 40.1\% &  61.5\% & 68.0\% \\
        GPT-2 0.1B + CF-3  & 11.9\% &  26.6\% & 32.9\% & 40.4\% &  61.0\% & 67.1\% \\
        GPT-2 0.1B + CF-2  & 11.1\% &  25.3\% &  31.7\% & 37.6\% &  57.8\% &  64.1\% \\
        GPT-2 0.1B + CF-1 & 10.0\% & 23.2\% & 28.8\% & 33.3\% &  54.1\% &  61.1\% \\
        \bottomrule
    \end{tabular}
    }
    \label{tab:results}
\end{table}

\subsection{RAG with Graph Textification}
\label{sec:eval-graph-rag}

We next compare \emph{ConceptFormer} to a text-based retrieval-augmented generation (RAG) approach. Specifically, for each subject entity, we retrieve its 1-hop neighbors from Wikidata and \emph{textify} them into a short passage appended to the LLM prompt \citep{talklikegraph, graphrag}. 

We use a simple template-based approach to convert a graph neighbourhood to text. The performance varied significantly depending on the injection template used, sometimes leading to over a 100\% difference in outcomes, consistent with findings from \cite{gptunderstands}. We observed that using a template in the form ``Subject (\{predicate\_1\}: \{object\_1\}, \{predicate\_2\}: \{object\_2\}, ...)'' performed particularly bad, while ``Subject, \{predicate\_1\} \{object\_1\}, \{predicate\_2\} \{object\_2\}, ...'', a template also used by \cite{textinjection}, performed best. However, this can easily span 100--800 tokens for well-known entities, consuming a significant portion of the LLM’s context window.

In Table~\ref{tab:results}, rows labeled ``+ RAG'' show large gains relative to the baseline. For smaller GPT-2 models, these gains can exceed $6\times$ at Hit@10. This underscores the potential of structured knowledge for IR if it is integrated effectively. However, the token overhead is substantial (on average 130 tokens per subject, but up to 800 for famous concepts), making it impractical in scenarios where multiple enriched entities appear in a single query. We also find that performance can degrade for large neighborhoods due to knowledge noise \citep{kbert}, in line with prior findings that excessive context can overwhelm the model.

\subsection{ConceptFormer Evaluation}
\label{sec:eval-conceptformer}

\emph{ConceptFormer} takes a vector-based approach to knowledge integration, drastically reducing the number of extra tokens needed for an entity’s subgraph. We instantiate \emph{ConceptFormer} with GPT-2 0.1B (125M parameters) to test whether a small model’s lack of knowledge can be compensated by structured prompts in \emph{concept-vector} format.
We employ the two-stage training scheme introduced in Section \ref{chapter:method}:
\begin{enumerate}
    \item \textbf{Pre-training on Tri-REx:} 
    We freeze GPT-2 0.1B and optimize \emph{ConceptFormer} to generate the correct next tokens for synthetic subject-predicate-object sentences. This fosters reliance on external subgraph data, since the model cannot simply recall them from memorization.
    \item \textbf{Fine-tuning on T-REx Bite:}
    We continue training \emph{ConceptFormer} on real Wikipedia sentences. This step ensures the approach generalizes to authentic textual contexts, not just the minimal triplets from the synthetic set.
\end{enumerate}

To explore the trade-off between vector capacity and prompt overhead, we train \emph{ConceptFormer} variants with $n \in \{1,2,3,4,5,10,15,20\}$. Each variant sees the same star subgraphs, but it can produce more or fewer concept vectors.

\begin{figure}
    \centering
    \includegraphics[width=1.0\linewidth]{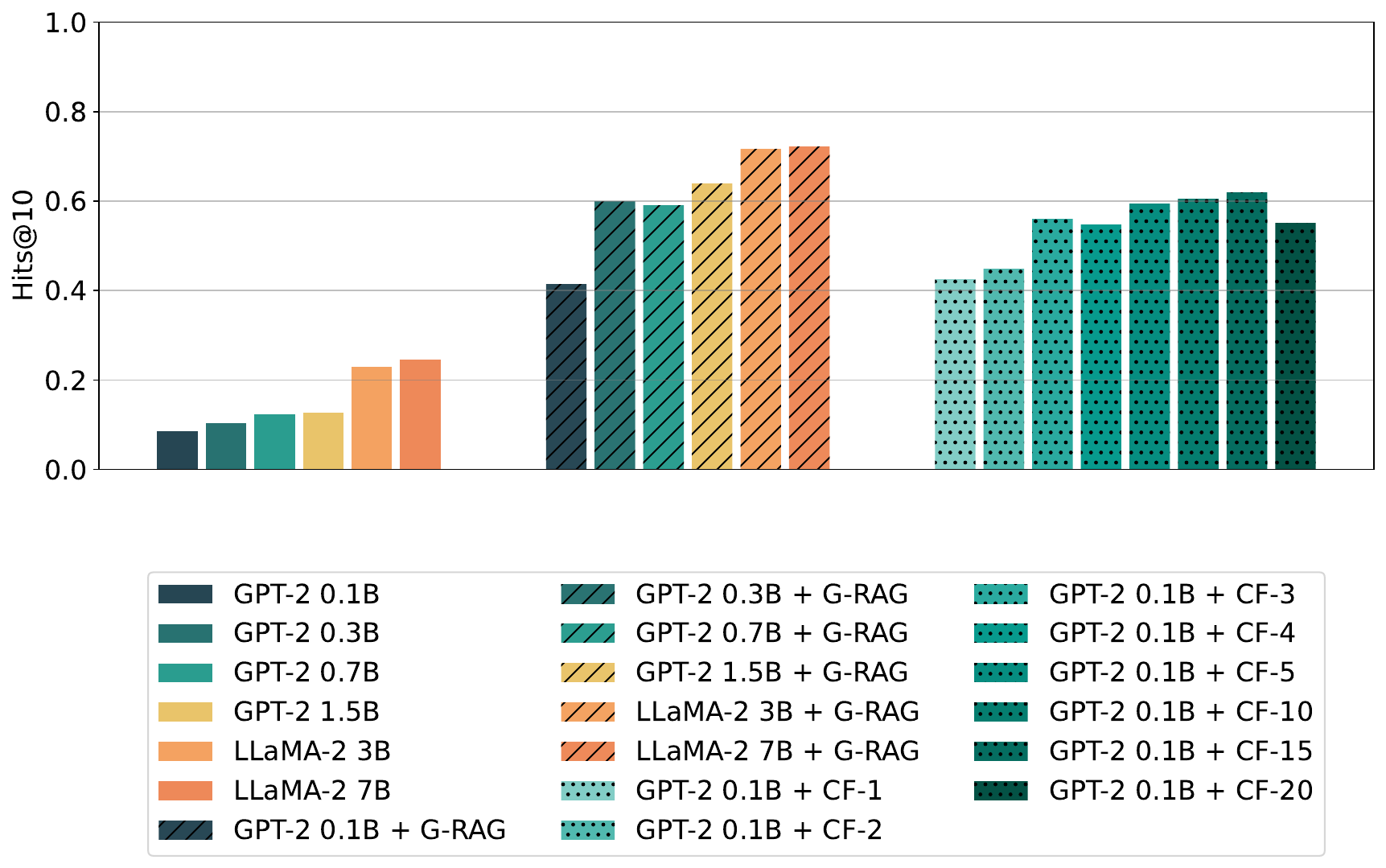}
    \caption{Hit@10 rate of various base models, with or without graph RAG (G-RAG), compared to GPT-2 0.1B with different \emph{ConceptFormers} (CF), after pre-training on \emph{Tri-REx}.}
    \Description{Hit@10 rate of various base models, with or without graph RAG (G-RAG), compared to GPT-2 0.1B with different ConceptFormers (CF), after pre-training on Tri-REx.}
    \label{fig:eval-sentenceformer-pretrained}
\end{figure}

Table~\ref{tab:results} (rows labeled ``GPT-2 0.1B + CF-$n$'') and Figure~\ref{fig:eval-sentenceformer-pretrained} illustrate that going from $n=1$ to $n=15$ markedly improves Hit@1 and Hit@10. Beyond 15, we see diminishing or no returns, suggesting that around 10--15 vectors are enough to cover typical 1-hop neighborhoods in Tri-REx. For GPT-2 0.1B, certain CF-$n$ models even outperform LLaMA-2 7B, a 50$\times$ larger model, in Hit@1—a striking result indicating that \emph{concept vectors} can encode essential knowledge more compactly than the model’s own parameters.

After fine-tuning on real sentences, Table~\ref{tab:results} shows the final performance. Notably, CF-15 yields a Hit@1 of 46.7\% and Hit@10 of 72.5\%---about a 10$\times$ gain over baseline GPT-2 0.1B. Even a single concept vector ($n=1$) outperforms text-based RAG for GPT-2 0.1B at Hit@1 (33.3\% vs.\ 6.6\%), while consuming \emph{130$\times$ fewer tokens} on average. 

\begin{figure}
    \centering
    \includegraphics[width=\linewidth]{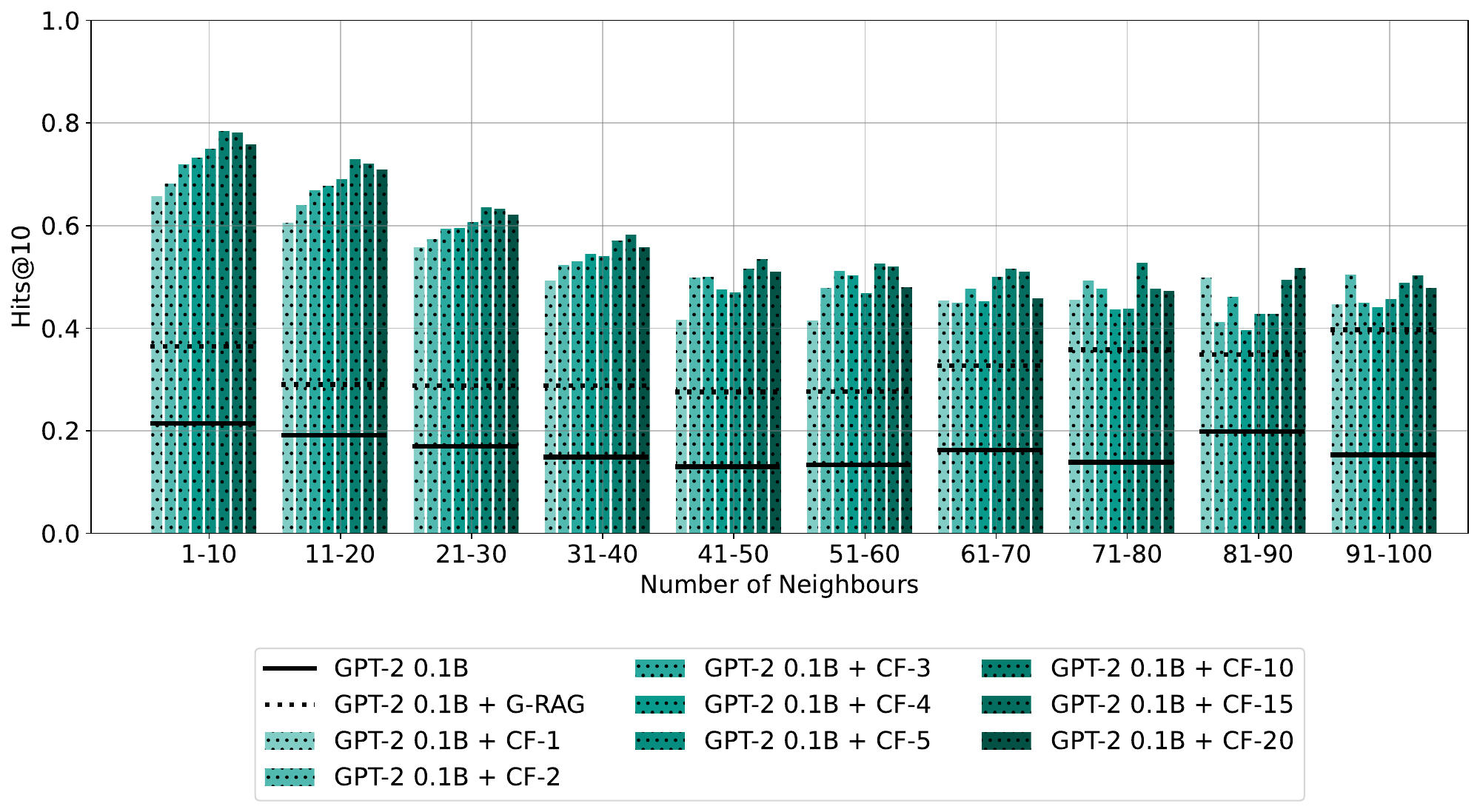}
    \caption{Hit@10 rate of GPT-2 0.1B + various \emph{ConceptFormers} on the Wikipedia based \emph{T-Rex Bite} Dataset.}
    \Description{Hit@10 rate of GPT-2 0.1B + various ConceptFormers on the Wikipedia based T-Rex Bite Dataset.}
    \label{fig:comparison-sentenceformers-trex-bite}
\end{figure}

Figure~\ref{fig:comparison-sentenceformers-trex-bite} highlights the trade-off: graph textification saturates the input context, particularly for well-known subjects, whereas \emph{ConceptFormer} retains high accuracy with minimal vector overhead. This is pivotal for IR scenarios where multiple enriched entities may appear simultaneously.

Following \citet{kbert}, we partition subjects into:
\begin{itemize}
    \item \textbf{Niche Concepts (1--10 neighbors):} Often less-known entities. CF-15 significantly outperforms RAG here, likely because text expansions for these entities are short, but the LLM still benefits from the learned concept representation.
    \item \textbf{Moderately Popular (11--90 neighbors):} Mixed results show consistent improvement for CF-15 over baseline GPT-2, though RAG also achieves good performance with moderate amounts of textual expansions.
    \item \textbf{Very Famous (90--100 neighbors):} Entities like ``Albert Einstein'' or ``Queen Elizabeth II'' can produce extremely large text expansions. RAG’s performance degrades, while CF-15 remains stable, illustrating that \emph{ConceptFormer} avoids knowledge noise issues when dealing with broad subgraphs.
\end{itemize}
Overall, \emph{ConceptFormer} maintains high recall across varying entity degrees, making it attractive for IR tasks spanning rare to well-known topics.

\subsection{Question Answering on WebQSP}
\label{sec:eval-webqsp}

To further validate \emph{ConceptFormer} in an IR-like question answering context, we evaluate it on the WebQuestions Semantic Parsing (WebQSP) dataset~\citep{wqsp}. Since WebQSP originally references Freebase, we use a Wikidata-linked variant \citep{webqspwikidata}. After filtering questions with missing or incompatible subgraphs, we form 2,463 question--answer pairs, each labeled with the relevant entity’s node ID.

We adapt GPT-2 0.1B to a prompt template: ``\texttt{Question: [Q]? Answer:}'', then measure whether the correct entity label emerges in the top-$k$ logits. As many questions have multiple valid answers or synonyms, we adopt a \emph{looser} Hit@5 threshold: if any correct label is in the top-5 tokens, we consider it a recall success.

Table~\ref{tab:webqsp-results} compares:
\begin{itemize}
    \item \textbf{Baseline GPT-2 0.1B}: 0\% Hit@1, effectively guessing randomly for these specialized QA prompts.
    \item \textbf{+ RAG (Graph Textification)}: Gains up to 1.4\% Hit@5 for GPT-2 0.1B, or 13.0\% for GPT-2 1.5B, but still low absolute performance. 
    \item \textbf{+ \emph{ConceptFormer}-10 (CF-10)}: Achieves 7.6\% Hit@1 and 28.3\% Hit@5, vastly surpassing the LLaMA-2 7B result of under 1\% Hit@1. 
\end{itemize}
While these numbers are modest compared to specialized QA systems that can reach 75\%+ \citep{ruijie}, they demonstrate that an off-the-shelf 0.1B-parameter LLM can become \emph{domain-aware} through \emph{ConceptFormer}’s vector-based injection. This highlights a path for using small LLMs in domain-oriented IR tasks (e.g., specialized KBs) without re-training a massive model.

\subsection{Additional Studies: Multi-Hop and Global Alignments}
\label{subsec:multi-hop-global}

Although our experiments focus on 1-hop subgraphs, \emph{ConceptFormer} can, in principle, be extended to multi-hop neighborhoods. Similarly, using globally aligned node embeddings (e.g., from PBG~\citep{pbg}) conferred no consistent advantage over simpler text-based embeddings. We hypothesize that \emph{ConceptFormer} itself internalizes enough global structure through repeated 1-hop exposures.

\subsection{Summary of Findings}

Across a spectrum of evaluation settings:
\begin{itemize}
    \item \textbf{Even a single \emph{concept vector}} ($n=1$) can boost GPT-2 0.1B from near-zero performance to competitive results while consuming $\sim100\times$ fewer tokens than text-based graph expansions.
    \item \textbf{15 \emph{concept vectors}} emerges as an effective upper bound for 1-hop neighborhoods, pushing recall higher but still remaining token-efficient.
    \item \textbf{Complex or Large Subgraphs} that hamper text-based RAG due to context saturation have less impact on \emph{ConceptFormer}, which compresses the subgraph into a fixed number of vectors.
    \item \textbf{General QA Scenarios} like WebQSP show that \emph{ConceptFormer} can transform a small LLM into a basic QA engine for domain knowledge queries. Performance, while not state-of-the-art, underscores the viability of vector-based knowledge injection.
\end{itemize}

Overall, our experiments confirm that \emph{ConceptFormer} provides a robust, scalable means to infuse structured knowledge into LLMs with minimal overhead—a boon for IR tasks where retrieving and presenting relevant factual data within tight context windows is critical. We next conclude by discussing limitations, future work, and broader implications for retrieval-augmented generation. 

\begin{table}[t]
    \centering
    \caption{Percentage of correctly answered questions from the WebQSP Dataset, comparison of the base models (BM), graph RAG (G-RAG), and \emph{ConceptFormer-10} (CF-10).}
    \resizebox{\linewidth}{!}{
    \begin{tabular}{lcccccc}
        \toprule
        & \multicolumn{2}{c}{\textbf{BM}} & \multicolumn{2}{c}{\textbf{G-RAG}} & \multicolumn{2}{c}{\textbf{CF-10}} \\
        \cmidrule(lr){2-3} \cmidrule(lr){4-5} \cmidrule(lr){6-7}
        \textbf{Model} & \textbf{H@1} & \textbf{H@5} & \textbf{H@1} & \textbf{H@5} & \textbf{H@1} & \textbf{H@5} \\
        \midrule
        LLaMA-2 7B & 0.9\% & 13.5\% & 0.1\% & 6.1\% &  &  \\
        LLaMA-2 3B & 0.1\% & 10.3\% & 0.0\% & 10.1\% &  &  \\
        GPT-2 1.5B & 0.1\% & 4.2\% & 0.2\% & 13.0\% &  &  \\
        GPT-2 0.7B & 0.0\% & 3.9\% & 0.1\% & 2.4\% &  &  \\
        GPT-2 0.3B & 0.0\% & 1.5\% & 0.0\% & 1.1\% &  &  \\
        GPT-2 0.1B & 0.0\% & 0.0\% & 0.2\% & 1.4\% & \underline{7.6\%} & \underline{28.3\%} \\
        \bottomrule
    \end{tabular}
    }
    \label{tab:webqsp-results}
\end{table}


\section{Conclusion} 
\label{chapter:conclusion}

This paper presented \emph{ConceptFormer}, a novel approach to augment Large Language Models with structured knowledge from Knowledge Graphs without modifying their internal architecture. Our experiments demonstrated that \emph{ConceptFormer} significantly enhances the factual recall abilities of GPT-2 0.1B. This improvement is attributed to the efficient transformation of KG information into compact and informative \emph{concept vectors}. The method of creating and injecting \emph{concept vectors} into the LLM input space offers a powerful way to enrich LLMs with current and detailed world knowledge while preserving token space for user queries or other contextual IR prompts.

From an IR perspective, \emph{ConceptFormer} provides a solution for knowledge-intensive tasks such as query expansion, entity-centric retrieval, and knowledge-grounded question answering. By embedding graph information at the input-embedding level rather than via verbose textual expansions, \emph{ConceptFormer} reduces the risk of context-window saturation—making it highly scalable for multi-entity queries often seen in advanced retrieval workflows. It bridges the gap between dense retrieval techniques and structured KG lookups, aligning with the broader shift toward retrieval-augmented generation methods in IR.

Our results show that \emph{ConceptFormer} achieves superior performance compared to raw LLMs and even outperforms template-based graph RAG methods in most scenarios (cf. Table~\ref{table:performance-change}). Notably, it enhances knowledge recall with a minimal increase in context size, providing an efficient pathway for integrating large-scale knowledge bases into modern IR pipelines. The effectiveness of \emph{ConceptFormer} is especially notable when comparing the input token usage between \emph{ConceptFormer-1} and graph RAG: just one singular \emph{concept vector} can outperform graph RAG on the \emph{T-Rex Bite} dataset by 88\% on Hit@10. Such savings directly benefit IR tasks that require injecting relevant knowledge for multiple entities within a single prompt, including complex queries or conversation-based searches.

Once trained, \emph{ConceptFormer} can pre-generate a comprehensive lookup table that maps entities to \emph{concept vectors}. Alternatively, it can be used dynamically, querying the relevant neighborhood from the source graph on the fly and embedding it into input space–compatible \emph{concept vectors}, thus offering a streamlined retrieval-augmentation step in IR pipelines. Changes in the online KG are automatically reflected and made accessible to the LLM, making \emph{ConceptFormer} suitable for highly dynamic retrieval scenarios with fast-changing graphs. Overall, it provides a versatile, token-efficient bridge between the structured world of KGs and the generative capabilities of LLMs, supporting more robust and up-to-date information retrieval.

\emph{ConceptFormer} exhibits four key properties that make it particularly compelling for IR pipelines:

\begin{description} 
    \item[Token Efficiency] Each entity’s neighborhood is compressed into a handful of vectors ($\approx$1--20 “soft tokens”), compared to hundreds of tokens required by typical text-based RAG expansions. IR scenarios often have limited context budgets, making such savings crucial for complex or multi-entity queries. 
    \item[No Fine-tuning of the LLM] By remaining purely in the input space, \emph{ConceptFormer} allows system integrators to reuse standard open-source or commercial LLMs. This is particularly valuable in contexts where re-training is infeasible, restricted for proprietary reasons, or prohibitively expensive. 
    \item[Adaptable to Any KG] The method is agnostic to the specific knowledge graph or embedding technique used. Any star topology subgraph can be fed in, allowing IR experts to integrate specialized domain knowledge (e.g., medical or legal) seamlessly into retrieval workflows. 
    \item[Scalability and Dynamic Updates] If the KG is large and relatively stable, precomputed \emph{concept vectors} can be quickly integrated. Conversely, if the KG is dynamic or requires real-time updates, on-the-fly generation remains feasible. This flexibility is especially advantageous for domains that demand constant knowledge updates. 
\end{description}

\begin{table}[t]
    \centering
    \caption{Performance change of GPT-2 0.1B + 15 \emph{concept vectors}, compared to GPT-2 0.1B base model and GPT-2 0.1B + graph RAG (G-RAG).}
    \resizebox{\linewidth}{!}{
    \begin{tabular}{rrrrrrr}
        \toprule
        & \multicolumn{3}{c}{\textbf{Tri-REx}} & \multicolumn{3}{c}{\textbf{T-Rex Bite}} \\
        \cmidrule(lr){2-4} \cmidrule(lr){5-7}
        \textbf{Model} & \textbf{H@1} & \textbf{H@5} & \textbf{H@10} & \textbf{H@1} & \textbf{H@5} & \textbf{H@10} \\
        \midrule
        GPT-2 0.1B  & 1121\% $\uparrow$ & 457\% $\uparrow$ & 348\% $\uparrow$ & 894\% $\uparrow$ & 370\% $\uparrow$ & 271\% $\uparrow$ \\
        GPT-2 0.1B + G-RAG  & 93\% $\uparrow$ & 0\% $\approx$ & 8\% $\downarrow$ & 612\% $\uparrow$ & 181\% $\uparrow$ & 123\% $\uparrow$ \\
        \bottomrule
    \end{tabular}
    }
    \label{table:performance-change}
\end{table}

These features underscore \emph{ConceptFormer}’s suitability for retrieval-based generation, question answering, or any IR task that demands up-to-date factual grounding. By prioritizing token efficiency, modularity, and dynamic scalability, \emph{ConceptFormer} fills a critical gap in bridging structured graph data with frozen Large Language Models while minimizing the burden on context budgets. Ultimately, it unifies structured graph knowledge with LLM-based generation in a manner that is practical, extensible, and highly aligned with the requirements of modern information retrieval systems.

\bibliographystyle{ACM-Reference-Format}
\bibliography{bibliography}

\end{document}